\documentclass{article} 
\usepackage{iclr2024_conference,times}

\usepackage{amsmath,amsfonts,bm}

\def\eqref#1{equation~\ref{#1}}

\def\1{\bm{1}}

\DeclareMathAlphabet{\mathsfit}{\encodingdefault}{\sfdefault}{m}{sl}
\SetMathAlphabet{\mathsfit}{bold}{\encodingdefault}{\sfdefault}{bx}{n}

\usepackage[hyperfootnotes=false]{hyperref}
\usepackage{url}
\usepackage{graphicx}
\usepackage{booktabs}
\usepackage{bbold}
\usepackage{cleveref}
\usepackage{siunitx}

\title{Towards DNA-Encoded Library Generation with GFlowNets}

\author{
Michał Koziarski\textsuperscript{*,1,2},
Mohammed Abukalam\textsuperscript{1,2},
Vedant Shah\textsuperscript{1,2},
Louis Vaillancourt\textsuperscript{2,3}, \\
\:\textbf{Doris Alexandra Schuetz\textsuperscript{2,3},
Moksh Jain\textsuperscript{1,2},
Almer van der Sloot\textsuperscript{1,2},
Mathieu Bourgey\textsuperscript{1,2},} \\
\:\textbf{Anne Marinier\textsuperscript{2,3},
Yoshua Bengio\textsuperscript{1,2}}
}

\newcommand\blfootnote[1]{%
    \bgroup
    \renewcommand\thefootnote{\fnsymbol{footnote}}%
    \renewcommand\thempfootnote{\fnsymbol{mpfootnote}}%
    \footnotetext[0]{#1}%
    \egroup
}

\iclrfinalcopy 
\begin{document}

\maketitle

\begin{abstract}
DNA-encoded libraries (DELs) are a powerful approach for rapidly screening large numbers of diverse compounds. One of the key challenges in using DELs is library design, which involves choosing the building blocks that will be combinatorially combined to produce the final library. In this paper we consider the task of protein-protein interaction (PPI) biased DEL design. To this end, we evaluate several machine learning algorithms on the PPI modulation task and use them as a reward for the proposed GFlowNet-based generative approach. We additionally investigate the possibility of using structural information about building blocks to design a hierarchical action space for the GFlowNet. The observed results indicate that GFlowNets are a promising approach for generating diverse combinatorial library candidates.
\end{abstract}

\blfootnote{
\textsuperscript{*} Corresponding author: \href{mailto:michal.koziarski@mila.quebec}{michal.koziarski@mila.quebec},
\textsuperscript{1} Mila,
\textsuperscript{2} Université  de Montréal,
\textsuperscript{3} IRIC}

\section{Introduction}

\begin{figure}[!htb]
    \centering
    \includegraphics[width=0.7\linewidth]{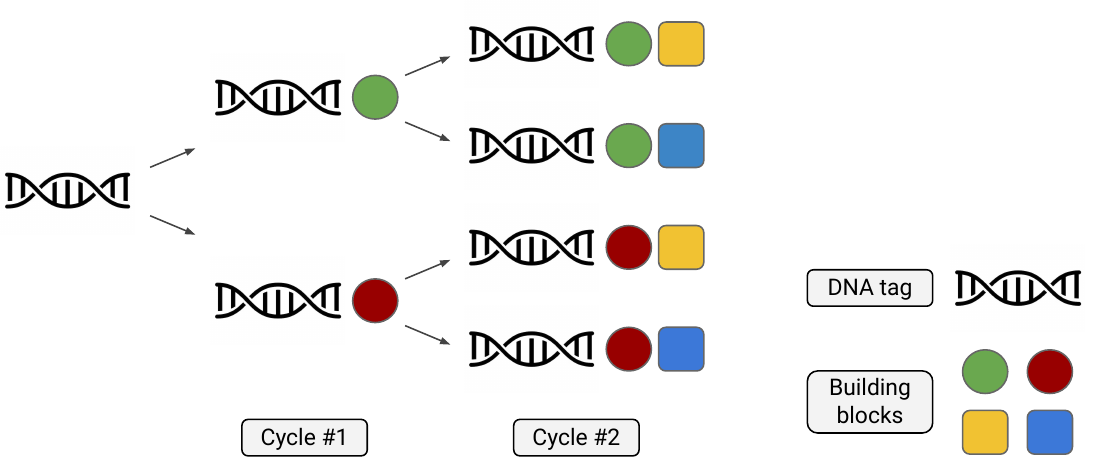}
    \caption{Illustration of DEL generation. Building blocks are first attached to a DNA tag used for identification, and later, across several consecutive cycles, combined in a combinatorial manner to produce a library of molecules built by joining together a sequence of building blocks.}
    \label{fig:del-abstract}
\end{figure}

Combinatorial libraries represent a powerful approach in chemistry for rapidly generating and screening large numbers of diverse compounds. These libraries are typically constructed by systematically combining building blocks or functional groups in various permutations. One example of combinatorial libraries is DNA-encoded libraries (DEL) \citep{goodnow2017dna,gironda2021dna,satz2022dna}. In DELs, unique DNA tags are attached to individual small molecule compounds, creating a vast library of DNA-tagged molecules (\Cref{fig:del-abstract}). Pulldown and next-generation sequencing (NGS) then allow for rapid identification of molecules with desired properties and potential drug candidates. While powerful, DELs are difficult to synthesize (the process can easily take several months), and might present challenges with respect to training machine learning models on the produced data \citep{mccloskey2020machine,lim2022machine}. Furthermore, a given set of building blocks may represent a large number of molecules resulting from their different permutations. This makes it difficult to evaluate a given set.

Due to the significant cost of synthesizing these libraries, it is desirable to construct versatile DELs, with high potential hit ratio (proportion of screened molecules with desired properties) across multiple targets, which are not known prior to the screen. This enables re-using the library in a multitude of screens. Moreover, biasing the libraries towards a specific family of targets can also be beneficial, as distinct regions of the chemical space are likely to produce higher hit rates for specific targets. One possible solution to fulfilling the above desiderata is the design of libraries biased towards modulating protein-protein interactions (PPIs) \citep{morelli2011chemical,bosc2020fr}. Dysfunctions in PPIs are known to be connected to various disease states. Despite this, PPI-targeting drugs are rare, mainly due to challenging ``druggability" \citep{morelli2011chemical} and poor quality of existing libraries. This makes the design of PPI-biased DELs potentially highly impactful.

In this paper we consider the task of computational DEL design. We utilize GFlowNets \citep{bengio2023gflownet} - a generative method recently used in multiple scientific discovery tasks \citep{bengio2021flow,jain2023gflownets,ai4science2023crystal,volokhova2023towards} due to its ability to produce highly diverse samples. We postulate that diversity will play an important role in DEL design, since it allows us to propose multiple library candidates with a wide range of chemical properties, out of which the best one can be determined experimentally. We first evaluate multiple machine learning algorithms in a PPI modulation classification task, and then use the best-performing method as a reward for the proposed GFlowNet. We demonstrate that the proposed approach achieves high PPI likelihood, while keeping the sample diversity high.

\section{Problem formulation}

In this paper we consider the problem of constructing DELs biased towards high likelihood of being PPI modulators. The motivation behind this is to develop general screening libraries that could eventually be used for targeting various proteins. Ultimately, we are interested in the problem of constructing very large ($>1$ million molecules) libraries, consisting of hundreds of building blocks selected out of thousands of viable ones, and possibly multiple chemical reactions assembling them. As a crucial stepping stone, in this paper we consider the simpler task of selecting a subset of already chosen building blocks, which translate to a specific library. This simplifies the problem, not only by significantly reducing the search space and required computational resources, but also by removing some of the biological constraints, for instance synthesizability of candidate building blocks.

\begin{figure}[!htb]
    \centering
    \includegraphics[width=0.9\linewidth]{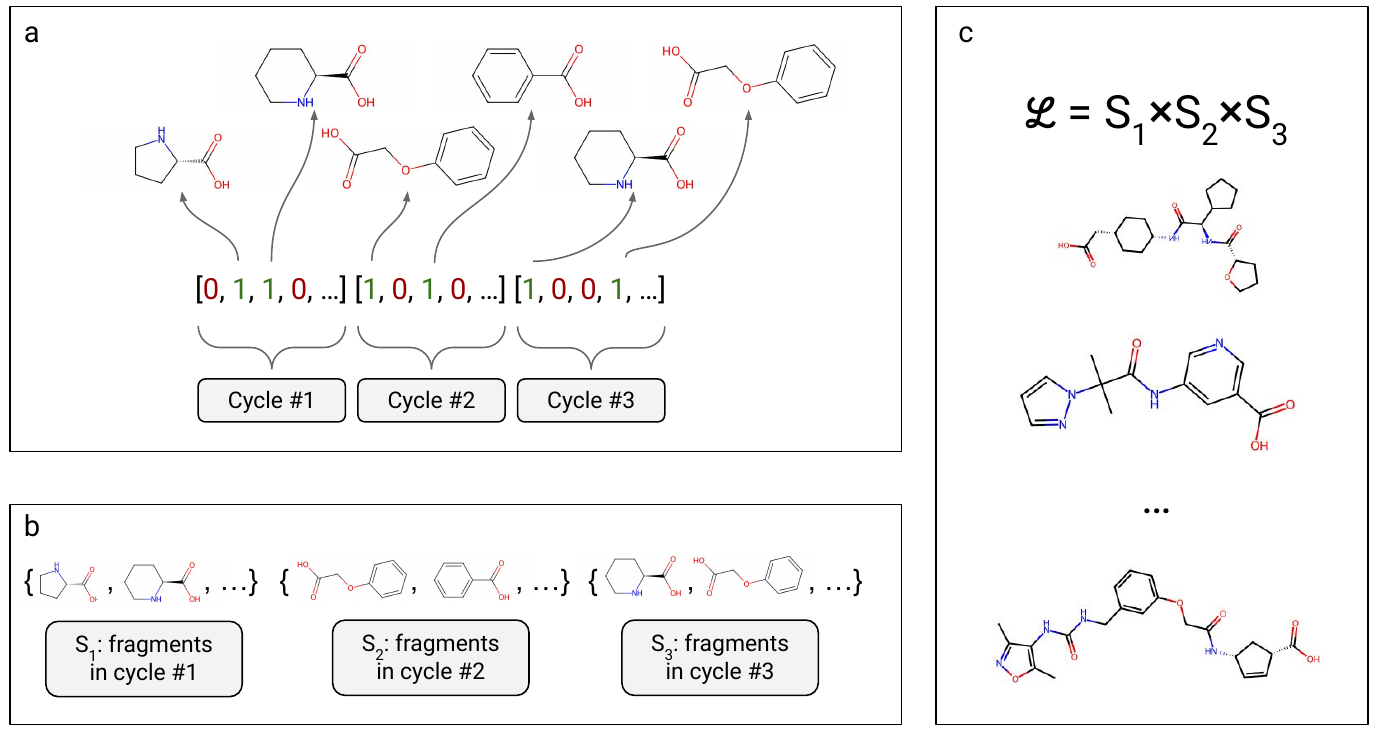}
    \caption{Illustration of the used state representation. a) Information about which building blocks are picked for a given cycle is represented as a binary vector, where each entry corresponds to a specific building block. b) For each cycle, a collection of selected building blocks is decoded. c) Resulting library is a Cartesian product of all the possible building blocks from different cycles, where each triplet of building blocks is combined to produce a final molecule.}
    \label{fig:problem-formulation}
\end{figure}

The way in which we represent states in presented in \Cref{fig:problem-formulation}. Specifically, the libraries we consider consist of three synthetic cycles, having 90, 89 and 197 possible building blocks, respectively, and a single way of combining the building blocks to generate a final molecule. This corresponds to a total of approximately 1.58 million molecules being present in the library, which corresponds to the Cartesian product of all the possible building blocks from different cycles. The task is finding specific sub-libraries (constructed by selecting subsets of these building blocks) that 1) produce a library of a specified size, 2) maximize the average likelihood of a molecule in the mentioned sub-library being a PPI modulator, and 3) the sub-libraries to be mutually diverse, which enables the users to choose particular ones depending on a specific target of interest. 
Point 3 incorporates the choice of libraries which tend to be biased towards specific chemical properties, such as molecular weight, number of rotatable bonds, polarity, etc.

We observe that the above problem can be stated as a binary vector search, with $i$-th element of the vector denoting whether $i$-th building block will be present in the corresponding library subset. More formally, we can represent a library by $x = x_1|x_2|x_3$, with $x_i$ denoting a sub-vector representing $i$-th cycle, and $x_1\in\{0, 1\}^{90}$, $x_2\in\{0, 1\}^{89}$ and $x_3\in\{0, 1\}^{197}$ (corresponding to the number of building blocks in different cycles). Then, if we denote the set of all available building blocks for $i$-th cycle as $\mathcal{B}_i$, and selected blocks as $\mathcal{S}_i = \{\mathcal{B}_{i,j}: \mathbb{1}[x_{i,j} = 1]\}$, the complete library subset that can be generated using $x$ is $\mathcal{L}(x) = \mathcal{S}_1 \times \mathcal{S}_2 \times \mathcal{S}_3$.

\section{DEL-GFlowNet}

\subsection{Flat variant}

GFlowNets \citep{bengio2021flow,bengio2023gflownet} are a family of generative methods designed to learn a sampling policy $\pi$ for constructing objects $x \in \mathcal{X}$ based on their non-negative reward function $R(x)$, such that $\pi(x) \propto R(x)$. This generation is done sequentially: starting from the initial state $s_0$, transitions $s_{t} {\rightarrow} s_{t+1} \in \mathbb{A}$ are applied between states $s \in \mathcal{S}$, forming trajectories $\tau=(s_0 \rightarrow s_1 \rightarrow \ldots \rightarrow x)$, where $\mathbb{A}$ is a predefined action space and $\mathcal{S}$ the state space. The policy $\pi$, e.g. a neural network, can then be optimized using one of the existing loss functions, such as Trajectory Balance \citep{malkin2022trajectory}.

We utilize the GFlowNet framework in DEL construction task by operating in the previously described state space $\mathcal{S}$ of binary vectors $x$ that can be mapped to corresponding libraries $\mathcal{L}(x)$, defining the action space $\mathbb{A} = \{1, 2, \dots, |x| + 1\}$, with action $a = i$ denoting that $i$-th bit of the vector $x$ should be flipped from 0 to 1 (plus one additional action denoting termination of sequence), and with a reward
\begin{equation}
    R(x) = \exp\left(\frac{\beta}{|\mathcal{L}(x)|}\sum_{i=1}^{|\mathcal{L}(x)|}{p(\mathcal{L}(x)_{i}})\right),
\end{equation}
with $p(m)$ denoting a function predicting the likelihood of the molecule $m$ representing a PPI modulator, and $\beta$ being a predefined parameter. We enforce the size constraints of the library by masking out the end of sequence action unless the minimum specified library size is reached, in addition to masking out every action that would increase the library size above the specified maximum. We refer to this approach as DEL-GFlowNet.

\subsection{Hierarchical variant}

Secondly, we observe that while the selected building blocks can be represented as a simple binary vector, this leads to loss of information associated with the specific molecular structure of the block (recall that building blocks themselves are small molecular fragments). Additionally, we note that in the basic variant, the action space of DEL-GFlowNet is relatively large (while to the best of our knowledge, there are no rigorous studies on the limits of GFlowNets, our own experience indicates that in problems with action spaces larger than one or two hundred, GFlowNets become difficult to train), and would have to be further enlarged to transition to bigger library sizes. This motivated us to consider the possibility of leveraging the structural information about building blocks to construct alternative, hierarchical action space that would be more compact.

\begin{figure}[!htb]
    \centering
    \includegraphics[width=0.9\linewidth]{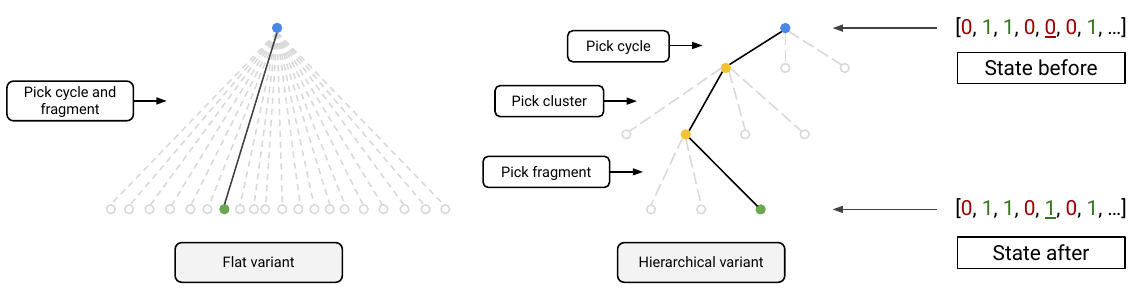}
    \caption{Illustration of action spaces of flat and hierarchical DEL-GFlowNet. Lines indicate actions (black: chosen, gray: other possible), and dots indicate states (blue: starting, yellow: intermediate, green: final). End effect in both cases is selection of one additional building block. Hierarchical variant allows us to significantly reduce the number of valid actions at any given stage.}
    \label{fig:hierarchical}
\end{figure}

To this end we first group the building blocks into clusters based on their molecular structure, and secondly split the action of picking a building block into three separate steps: 1) picking the cycle, 2) picking one of the clusters for the given cycle, and 3) picking one of the building blocks belonging to that cluster (\Cref{fig:hierarchical}). Additionally, to make the policy learning feasible, we augment the state space with one-hot encodings of the current cycle and cluster, as well as two additional binary values indicating whether any cycle and cluster were already picked. We perform clustering using Agglomerative Clustering on 2048-dimensional ECFPs \citep{rogers2010extended}, with Jaccard similarity as a distance measure, setting the number of clusters to 10 for cycles 1 and 2, and to 20 for cycle 3. We refer to this hierarchical variant as H-DEL-GFlowNet.

\section{Experimental study}

\subsection{Proxy training}

Since the reward function is an essential component of the GFlowNet training, we begin by examining the possibility of training a proxy model $p(m)$ capable of reliable prediction of PPI modulation likelihood. To this end we construct a dataset consisting of 2583 compounds from the 2P2Idb database \citep{basse20122p2idb} with confirmed orthosteric binding to the target (treated as positives) and 1541 FDA approved drugs (treated as negatives). Note that we exclude FDA approved drugs known to modulate PPI. We make a key assumption that remaining known drugs are unlikely to modulate PPI, since that would lower their specificity. We divide this dataset into train, validation and test partitions using scaffold splitting into a 0.8:0.1:0.1 proportion.

We evaluate the performance of several classification algorithms: Random Forest (RF) trained on 2048-dimensional ECFPs, a pretrained Molformer \citep{wu2023molformer}, and a graph neural network (GNN) - both trained from scratch and pretrained. The details of the training can be found in \Cref{apx:proxy}. The results of this comparison were presented in \Cref{tab:proxy}. As can be seen, the best performance was achieved by a pretrained GNN. While neural network-based methods outperformed RF by a margin, in general all of the classification algorithms achieved reasonable performance, indicating that they would perform well in the PPI modulator discrimination.

\begin{table}[!htb]
    \caption{Comparison of different classification algorithms for PPI modulation prediction. Best performance denoted in bold.}
    \label{tab:proxy}
    \centering
    \begin{tabular}{lcccccc}
        \toprule
        Method & Accuracy & Precision & Recall & AUC & F1-score & AP \\
        \midrule
        RF & 0.748 & 0.707 & \textbf{0.962} & 0.865 & 0.815 & 0.893 \\
        Molformer (pretrained) & 0.799 & 0.819 & 0.836 & 0.872 & 0.827 & 0.898 \\
        GNN & 0.818 & 0.856 & 0.824 & 0.887 & 0.839 & 0.907 \\
        GNN (pretrained) & \textbf{0.823} & \textbf{0.857} & 0.832 & \textbf{0.898} & \textbf{0.844} & \textbf{0.917} \\
        \bottomrule
    \end{tabular}
\end{table}

\begin{figure}[!htb]
    \centering
    \includegraphics[width=0.9\linewidth]{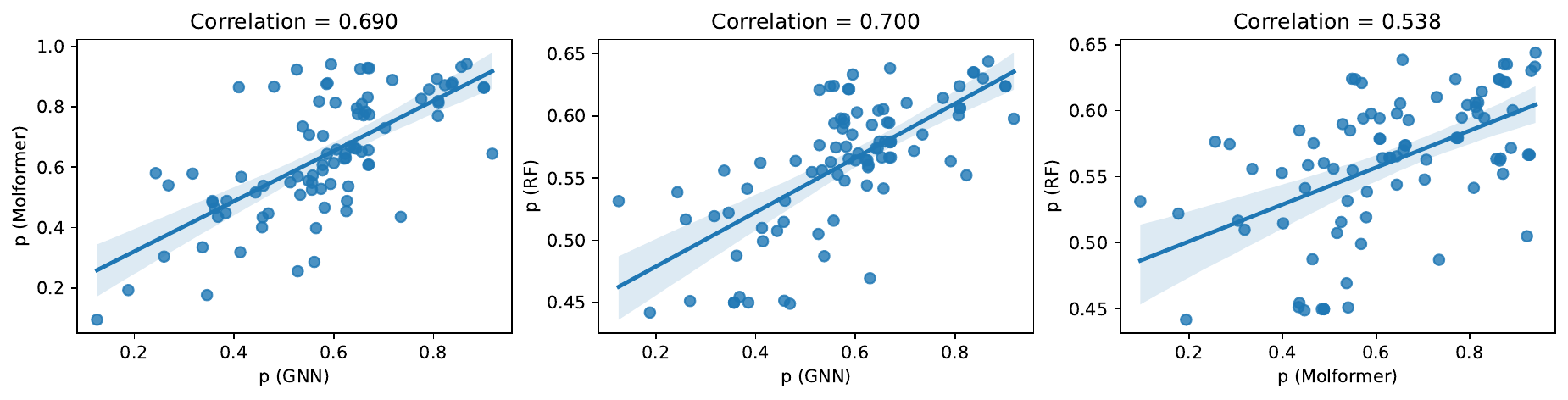}
    \caption{Comparison of model similarity. Individual points represent specific building blocks from first cycle, with average probability values of molecules containing that building block.}
    \label{fig:corr}
\end{figure}

This claim was further tested in an additional experiment, in which we compared the individual quality of building blocks belonging to one of the cycles. We did so by comparing average probabilities of molecules containing given building block being a PPI modulator, outputted by different models. The results were presented in \Cref{fig:corr}. As can be seen, while some correlation between the models was observed, it was not perfect. Our conclusion is that while all of the models achieved comparable results on the small dataset of known PPI modulators, this only partially translates to the results on the whole library, making the choice of the optimal model difficult in practice.

\subsection{Library sampling}
\label{sec:sampling}

Using the best performing model on the dataset of known PPI modulators (pretrained GNN) as the reward model, we evaluate the proposed DEL-GFlowNet method in a library subset sampling task. We compare it with four baselines: random sampling, a greedy approach in which 25/25/40 individually best performing blocks were selected per cycle, Markov chain Monte Carlo (MCMC) \citep{robert2004metropolis}, and proximal policy optimization (PPO) \citep{schulman2017proximal}. We consider the task of sampling a library with a size between 20k and 25k molecules. Note that partial results for sampling a 90k to 100k molecules library can be found in \Cref{apx:large}. Training details can be found in \Cref{apx:training-details}. In total, we sample 5,000 samples for each method, and averaged the results across 3 random runs. We report both the average PPI likelihood of the sampled libraries, as well as their diversity:
\begin{equation}
\operatorname{Diversity}(\mathcal{D})=\frac{\sum_{\left(x_{i}, y_{i}\right) \in \mathcal{D}} \sum_{\left(x_{j}, y_{j}\right) \in \mathcal{D} \backslash\left\{\left(x_{i}, y_{i}\right)\right\}} d\left(x_{i}, x_{j}\right)}{|\mathcal{D}|(|\mathcal{D}|-1)},
\end{equation}
directly on the binary vectors, with hamming distance used as distance measure. 

\begin{figure}[!htb]
    \centering
    \includegraphics[width=\linewidth]{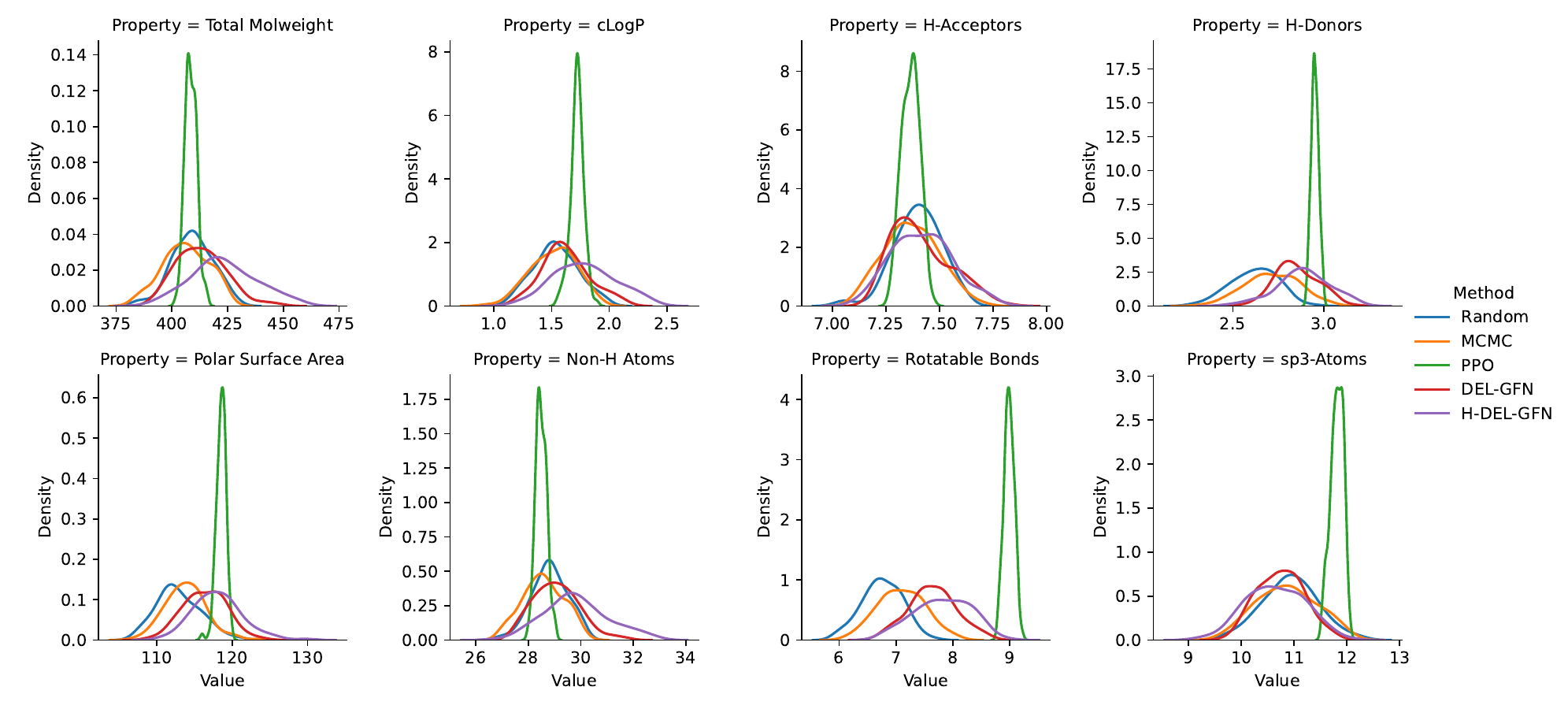}
    \caption{Distributions of average chemical library properties of top-100 generated libraries.}
    \label{fig:dist-25k}
\end{figure}

We report the numerical results in \Cref{tab:sampling}. We compute both distance and average probability in three settings: across all samples, for top-100 candidate libraries with highest reward, and for the best sampled candidate. Additionally, to further illustrate the diversity, in \Cref{fig:dist-25k} we visualize the distributions of average values of several chemical properties of the generated libraries. Several observations can be made based on the results. First of all, it's worth noting that the simple greedy approach outperforms all of the other methods in a single library design, finding a library candidate with near-perfect probability. However, we argue that taking into account the uncertainty associated with the quality of the proxy, in practice it is unlikely to produce the "optimal" (with respect to real, unknown proportion of possible PPI modulators) library. Instead, what is required is proposal of multiple diverse library candidates which can be evaluated experimentally, which is something the greedy approach is incapable of. PPO behaves similarly: while it achieves very good performance of an individual library, it collapses to a single mode, and has low diversity of solutions. GFlowNets outperform remaining methods, producing libraries with higher estimated PPI likelihood and slightly higher diversity. The second point is further illustrated in \Cref{fig:dist-25k}, where we can observe GFlowNet-produced libraries to have wider ranges of selected average properties, including molecular weight, cLogP and the number of non-hydrogen atoms. Finally, we note that this is more pronounced for H-DEL-GFlowNet than DEL-GFlowNet, which also achieves slightly higher performance as shown in \Cref{tab:sampling}.

\begin{table}[!htb]
    \caption{Comparison of different DEL sampling strategies. Best performance denoted in bold.}
    \label{tab:sampling}
    \centering
    \begin{tabular}{lccccc}
        \toprule
        Method & Prob. & Div. & Top-100 prob. & Top-100 div. & Top-1 prob. \\
        \midrule
        Random & 0.582 ± 0.000 & 0.378 ± 0.000 & 0.706 ± 0.002 & 0.374 ± 0.001 & 0.754 ± 0.011 \\
        Greedy & - & - & - & - & \textbf{0.992} \\
        MCMC & 0.696 ± 0.000 & 0.393 ± 0.000 & 0.809 ± 0.000 & 0.385 ± 0.001 & 0.850 ± 0.002 \\
        PPO & \textbf{0.974 ± 0.006} & 0.105 ± 0.015 & \textbf{0.983 ± 0.005} & 0.090 ± 0.012 & 0.985 ± 0.004 \\
        DEL-GFN & 0.767 ± 0.003 & 0.397 ± 0.001 & 0.873 ± 0.003 & 0.383 ± 0.001 & 0.912 ± 0.011 \\
        H-DEL-GFN & 0.783 ± 0.013 & \textbf{0.402 ± 0.001} & 0.885 ± 0.010 & \textbf{0.390 ± 0.001} & 0.918 ± 0.008 \\
        \bottomrule
    \end{tabular}
\end{table}

\section{Discussion \& future directions}

In this paper we frame the combinatorial library design problem as a binary vector search task and design DEL-GFlowNet - a generative method for sampling diverse library candidates biased towards PPI modulators. Furthermore, we extended the approach by introducing a hierarchical action space based on clustering the molecular structures of the building blocks. We evaluate the approach on a task of selecting a subset of existing library. While the results seem promising, as demonstrated by the ability to sample high-quality libraries with diverse chemical properties, there are also several limitations that would likely need to be addressed in order to scale the method to facilitate the design of significantly larger libraries.

One challenge lies in training the classification algorithm predicting the likelihood of a molecule being a PPI modulator. While we were able to train multiple classification algorithms, each achieving good performance on a small dataset of molecules with known PPI modulation activity, and their performance is roughly comparable on that dataset, non-negligible differences were observed with respect to ranking individual building blocks of the considered DEL. This indicates that the performance on the small dataset of known PPI modulators does not necessarily translate to reliable out-of-distribution prediction on DEL, and poses additional difficulty related to ambiguity of which proxy model should be used as a reward for GFlowNet training. While typically this problem can be solved by gathering additional data, in the case of general PPI modulation this becomes tricky. This is because while positives can be identified somehow reliably, this is not the case for negatives (while a molecule can be labeled a PPI modulator if it modulates at least one PPI, determining if it will not be able to modulate any PPI is difficult).

Secondly, while we observed promising results for sampling library candidates with GFlowNets, several issues remain here as well. Most importantly, the considered problem of library subset selection is still relatively small, and it is not clear how the performance would scale with a significantly increased number of possible building blocks and/or reactions combining them. This would also introduce an additional consideration: since for the proposed DEL the total possible number of generated molecules was manageable (1.58 million), making it feasible to just precompute proxy probabilities for every molecule, larger variants might require e.g. stochastically sampling smaller batches of possible rewards for the sake of reward computation. Finally, while some (quite modest) improvement in performance was observed for the hierarchical approach for the smaller library subset, it wasn't observed in the case of a larger subset, making it inconclusive to what extent introducing hierarchy is beneficial. Nevertheless, we argue that since the approach itself is quite naive, in particular due to the use of ECFPs on small molecular fragments for clustering, further investigation might be warranted.

\section*{Acknowledgment}

We would like to express our gratitude to the CIFAR (Canadian Institute for Advanced Research), Consortium Acuité Québec, FACS (Fonds d’accélération des collaborations en santé), IVADO/PRF3, Genentech and Genome Quebec for their generous support in funding this research project.

\bibliography{iclr2024_conference}
\bibliographystyle{iclr2024_conference}

\appendix

\section{Training details}
\label{apx:training-details}

\subsection{Proxy}
\label{apx:proxy}

\textbf{GNN} model consisted of 5 GIN layers \cite{xu2018powerful} with hidden dimensionality of 500, utilized Jumping Knowledge shortcuts \cite{xu2018representation} and had a single output MLP layer. Pretraining was done in an unsupervised fashion on ZINC15 dataset \cite{sterling2015zinc}. Training was done for 30 epochs using Adam optimizer with learning rate of \num{5e-5} and batch size of 50. \textbf{Molformer} used published pretrained weights, and had 12 layers with dimensionality of 768 each. Finetuning was done for 50 epochs, using learning rate of \num{3e-5} and batch size of 16. \textbf{Random Forest} used 1000 trees.

\subsection{GFlowNet}
\label{apx:gflownet}

GFlowNets were trained for 5000 training iterations, using learning rate of \num{1e-4}, batch size of 50 forwards samples and 50 replay samples, with prioritized replay buffer of up to 1000 samples. Both forward and backward policies were 5 layer MLPs, with hidden dimensionality of 512. Training was done using Trajectory Balance loss, $\beta = 64$ and random action probability of 0.1.

\subsection{MCMC}

We used Metropolis–Hastings variant of MCMC. We initialize chains by sampling random binary vectors corresponding to the library size within specified size constraints. Importantly, unlike for GFlowNets, we allow bidirectional bit flipping. We use the number of random chains equal to number of target samples (5000), and use chain length of 250. Importantly, we disallow actions that would lead to exceeding specified library size, and do not count them towards the chain length. Note that in this setting, we use $5\times$ the budget of proxy calls compared to the GFlowNet.

\subsection{PPO}

The PPO policy was trained for 2000 iterations. The training converges within 1500 iterations. We use a learning rate of \num{1e-4} with a decay period of \num{1e6} and a decay coefficient of 0.5. We use Adam optimizer, a batch size of 2, a random action probability of 0.001 and a reward scaling factor ($\beta$) of 64. The policy is a MLP with two hidden layers of size 256 each. We use a clipping factor ($\epsilon$) of 0.1 and an entropy coefficient of \num{1e-3}. We collect 64 trajectories per iteration and train for 16 epochs per iteration.

\section{Large library sampling}
\label{apx:large}

The experiments from \Cref{sec:sampling} were repeated with library size constrained between 90k and 100k. Note that in contrast to the experiment with smaller library, due to computational considerations batch size was restricted to 25 forward samples and 25 replay samples. For the greedy baseline, we used 35/35/80 building blocks per cycle. Importantly, the large scale experiment did not seem to converge in the specified number of iterations, so the results should be treated as preliminary.

\begin{table}[!htb]
    \caption{Comparison of different DEL sampling strategies, with requested library size between 90k and 100k. Best performance denoted in bold.}
    \label{tab:sampling-100k}
    \centering
    \begin{tabular}{lccccc}
        \toprule
        Method & Prob. & Div. & Top-100 prob. & Top-100 div. & Top-1 prob. \\
        \midrule
        Random & 0.582 ± 0.000 & \textbf{0.480 ± 0.000} & 0.671 ± 0.002 & \textbf{0.473 ± 0.001} & 0.726 ± 0.013 \\
        Greedy & - & - & - & - & 0.975 \\
        MCMC & 0.619 ± 0.000 & 0.478 ± 0.000 & 0.706 ± 0.002 & 0.466 ± 0.001 & 0.747 ± 0.012 \\
        PPO & \textbf{0.977 ± 0.003} & 0.103 ± 0.010 & \textbf{0.986 ± 0.002} & 0.086 ± 0.008 & \textbf{0.988 ± 0.001} \\
        DEL-GFN & 0.669 ± 0.001 & 0.475 ± 0.000 & 0.755 ± 0.002 & 0.455 ± 0.001 & 0.797 ± 0.003 \\
        H-DEL-GFN & 0.656 ± 0.004 & 0.476 ± 0.001 & 0.749 ± 0.003 & 0.456 ± 0.001 & 0.786 ± 0.004 \\
        \bottomrule
    \end{tabular}
\end{table}

\begin{figure}[!htb]
    \centering
    \includegraphics[width=\linewidth]{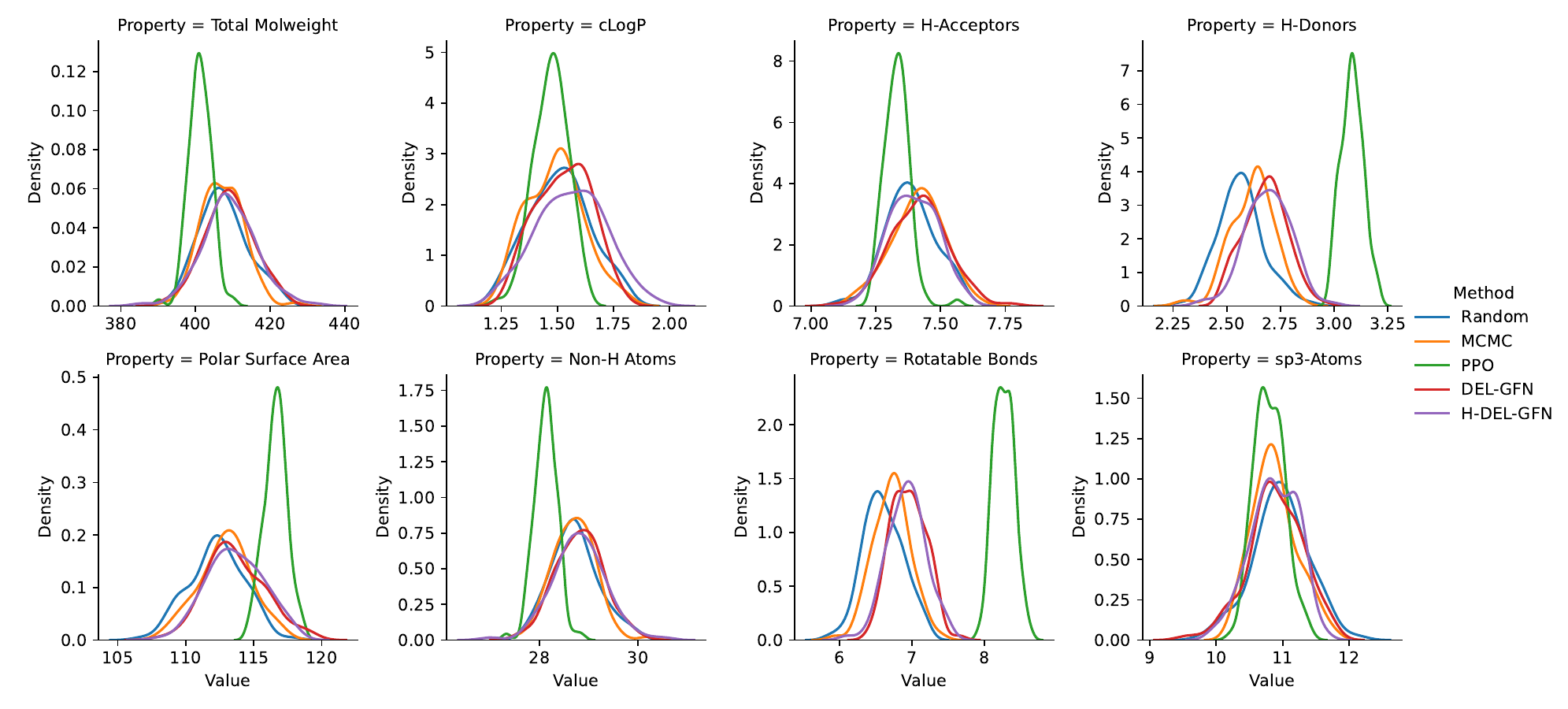}
    \caption{Distributions of average chemical library properties of top-100 generated libraries, with requested library size between 90k and 100k.}
    \label{fig:dist-100k}
\end{figure}

\end{document}